\documentclass[a4paper,twoside]{article}

\usepackage{epsfig}
\usepackage{subcaption}
\usepackage{calc}
\usepackage{amssymb}
\usepackage{amstext}
\usepackage{amsmath}
\usepackage{amsthm}
\usepackage{multicol}
\usepackage{pslatex}
\usepackage{apalike}
\usepackage{algorithm2e}
\usepackage[bottom]{footmisc}

\usepackage{graphicx}
\usepackage{booktabs}
\usepackage{hyperref}
\usepackage{array}
\usepackage{float} 
\usepackage{color}
\usepackage{multirow}

\usepackage{SCITEPRESS}     

\begin{document}

\title{Local Foreground Selection aware Attentive Feature Reconstruction for few-shot fine-grained plant species classification}

\author{\authorname{Aisha Zulfiqar\sup{1}, Ebroul Izquiedro\sup{1}}
\affiliation{\sup{1} Queen Mary University of London, London, United Kingdom}
}


\abstract{Plant species exhibit significant intra-class variation and minimal inter-class variation. To enhance classification accuracy, it is essential to reduce intra-class variation while maximizing inter-class variation. This paper addresses plant species classification using a limited number of labelled samples and introduces a novel Local Foreground Selection(LFS) attention mechanism. LFS is a straightforward module designed to generate discriminative support and query feature maps. It operates by integrating two types of attention: local attention, which captures local spatial details to enhance feature discrimination and increase inter-class differentiation, and foreground selection attention, which emphasizes the foreground plant object while mitigating background interference. By focusing on the foreground, the query and support features selectively highlight relevant feature sequences and disregard less significant background sequences, thereby reducing intra-class differences. Experimental results from three plant species datasets demonstrate the effectiveness of the proposed LFS attention mechanism and its complementary advantages over previous feature reconstruction methods.}

\onecolumn \maketitle \normalsize \setcounter{footnote}{0} \vfill

\section{\uppercase{Introduction}}
\label{sec:introduction}

Automatic plant species classification is essential for biodiversity conservation, as understanding the plant species present in a habitat is crucial for its preservation. This classification task is a fine-grained image classification problem, characterized by subtle differences between species. Large intra-class variations arise from differences in background, illumination, and the pose of various plant organs, while inter-class variations tend to be minimal. Consequently, effective training must focus on learning discriminative features to enhance classification performance.

Few-shot image classification \cite{tang2020revisiting} \cite{tsutsui2019meta}  is particularly challenging due to limited training data. In fine-grained contexts, it involves classifying using only a few images per class, necessitating the optimization of both inter-class and intra-class variations. Conventional few-shot learning  \cite{NIPS2017} \cite{sung2018learning} typically addresses only class-level differences, which limits its effectiveness for fine-grained classification tasks.

Recently, innovative approaches have emerged for fine-grained image classification that leverage feature reconstruction \cite{wertheimer2021few} \cite{doersch2020crosstransformers}. In these methods, support image features reconstruct query features to predict their classes, either through uni-directional reconstruction \cite{wertheimer2021few} or bi-directional reconstruction \cite{wu2023bi}, thereby decreasing intra-class variations and increasing inter-class variations. A significant source of intra-class differences in fine-grained objects is the background present in the images, indicating further opportunities for optimizing intra-class differences.

In this study, we present a novel Local Foreground Selection (LFS) attention mechanism, designed to localize discriminative regions through a dual-action attention approach that generates distinct features. The LFS attention mechanism effectively reduces background effects while emphasizing the foreground, along with capturing essential local discriminative details. This is achieved by combining local attention, which extracts local spatial context to enhance inter-class variation, and foreground selection attention, which minimizes intra-class variation by highlighting foreground object and reducing background effects. The novelty of the LFS attention mechanism lies in its integration of local and foreground selection attention, which together outperform their individual applications. This attention mechanism produces highly discriminative features, facilitating the reconstruction of support and query features in existing few-shot fine-grained classification methods  \cite{wertheimer2021few} \cite{wu2023bi}. Our main contributions can be summarized as follows:

\begin{itemize}
	\item In this study, we propose a novel dual-action Local Foreground Selection(LFS) attention mechanism designed to optimize both inter-class and intra-class variations. Our proposed attention generates discriminative feature maps. When combined with existing feature reconstruction approaches for few-shot finegrained classification LFS achieves state-of-the-art performance. 
	\item Our work focuses exclusively on plant species classification using three plant species datasets. The datasets contain images with natural backgrounds, which present significant classification challenges.
\end{itemize}

\section{\uppercase{Related Works}}

\subsection{Metric based few shot learning}

Among meta-learning techniques, metric learning is the most widely employed approach. Common metric-based methods in few-shot learning include Matching Network \cite{vinyals2016matching} , Prototypical Network \cite{NIPS2017} , and Relation Network \cite{sung2018learning}. In these methods, training data is converted into feature vectors located within a feature space corresponding to their classes. The similarity between two feature vectors is quantified by the distance between them; a smaller distance indicates greater similarity. The class of a query sample is determined by calculating its distance to feature vectors of different classes, with the class corresponding to the closest feature vector being assigned.

Typically, distance metrics used include Euclidean distance or cosine distance, as well as other learned parametric distance metrics. Metric-based methods are favoured for their simplicity and efficiency. Prior approaches specifically for fine-grained images include a low-rank pairwise bilinear pooling method designed to learn an effective distance metric \cite{huang2020low}, and a focus area location mechanism to identify similar regions among objects \cite{sun2020few}. The multi-attention meta-learning (MattML) method employs a metric approach to address the fine-grained classification problem \cite{zhu2020multi}. Additionally, a local descriptor-based image-to-class measure technique \cite{li2019revisiting} learns feature metrics, while a similarity measure method proposed by a non-linear data projection network enhances fine-grained image classification \cite{zhang2021ndpnet}. The bi-similarity network \cite{li2020bsnet} employs a dual similarity check to obtain more discriminative features, and the target-oriented alignment network \cite{huang2021toan} matches support and query images to reduce intra-class variance while increasing inter-class variance.

Common convolutional feature extractors such as ConvNet, ResNet, and WRN are utilized to learn the metric spaces. These feature extractors generate feature maps that characterize appearance across a grid of spatial locations. However, the selected distance functions typically require a single vector representation for the entire image. Ideally, this conversion should preserve spatial granularity and detail from the feature map without over fitting to pose, but current methods fall short of achieving this. Consequently, metric-based methods often under perform in fine-grained image classification due to their inability to account for the high similarities between categories.

\subsection{Feature reconstruction method}
While metric-based methods necessitate the creation of a single vector that preserves spatial locations, the feature reconstruction approach addresses this limitation. DeepEMD \cite{zhang2020deepemd} does not perform matching at the image level like traditional metric learning methods; instead, it partitions the image into a set of local representations. Optimal matching is conducted on these representations from two images to assess similarity, with Earth Mover's Distance employed to compute the distance between them. However, the DeepEMD approach involves a high computational cost for reconstruction.

In the Feature Reconstruction Network (FRN) \cite{wertheimer2021few}, feature maps are reconstructed by pooling support set feature maps into a matrix, where each column represents the concatenated feature maps of a channel. For classification, each location in the query image's feature map is reconstructed using a weighted sum of the support features from the corresponding class. The LCCRN \cite{li2023locally} network enhances local information extraction by introducing a local content extraction module, while a separate embedding module preserves appearance details. LCCRN simultaneously learns both appearance and local details, resulting in four distinct reconstruction tasks.

Bi-FRN \cite{wu2023bi} adopts a feature reconstruction strategy; however, unlike previous methods, it reconstructs query images from support images and vice versa. This results in a feature pool derived from both support and query images, with pooled features mutually reconstructed from one another. In the context of fine-grained ship classification \cite{li2022few}, foreground weights of query feature maps are generated using both parametric and non-parametric weight generators, with the distance metric derived from the proposed reconstruction error of query features calculated using weighted squared Euclidean distance.

\subsection{Attention Mechanism}
The transformer self-attention mechanism \cite{vaswani2017attention} has been incorporated into several few-shot learning methods. The Few-shot Embedding Adaptation with Transformer (FEAT) \cite{ye2020few} utilizes the self-attention mechanism of the Transformer to perform task-specific embedding adaptation. This approach employs a set-to-set function to derive more discriminative instance representations and models interactions among images within each set. 

The CTX model \cite{doersch2020crosstransformers} proposes a self-supervised learning framework combined with a Cross Transformer, representing images as spatial tensors and generating query-aligned prototypes. CTX leveraged self-attention to identify the spatial attention weights of support and query images, facilitating the learning of query-aligned class prototypes.

Building on the attention mechanism of transformers, the Few-shot Cosine Transformer (FS-CT) \cite{nguyen2023enhancing} introduces cosine attention, which has been shown to outperform standard softmax attention in few-shot tasks. This proposed cosine attention results in a more effective correlation map between support and query images, leading to improved predictions for the class of the query image. Additionally, the Bi-directional Feature Reconstruction Network (Bi-FRN) \cite{wu2023bi} utilizes a self-attention mechanism to generate discriminative query and support features.

\section{\uppercase{Method}}
\label{sec:method}

Classification of plant species represents a fine-grained image classification task, characterized by subtle differences among various species. This challenge is compounded by large intra-class variations and minimal inter-class differences that must be addressed. Additionally, plant classification is conducted within their natural environments, necessitating the removal of background effects.

We propose a novel attention-based module designed to generate discriminative feature maps for use with existing few-shot fine-grained approaches for plant species classification. This novel module can enhance the performance of feature reconstruction approaches demonstrated in prior works \cite{wu2023bi} \cite{wertheimer2021few}.

\begin{figure}[!h]
  \centering
   {\epsfig{file = 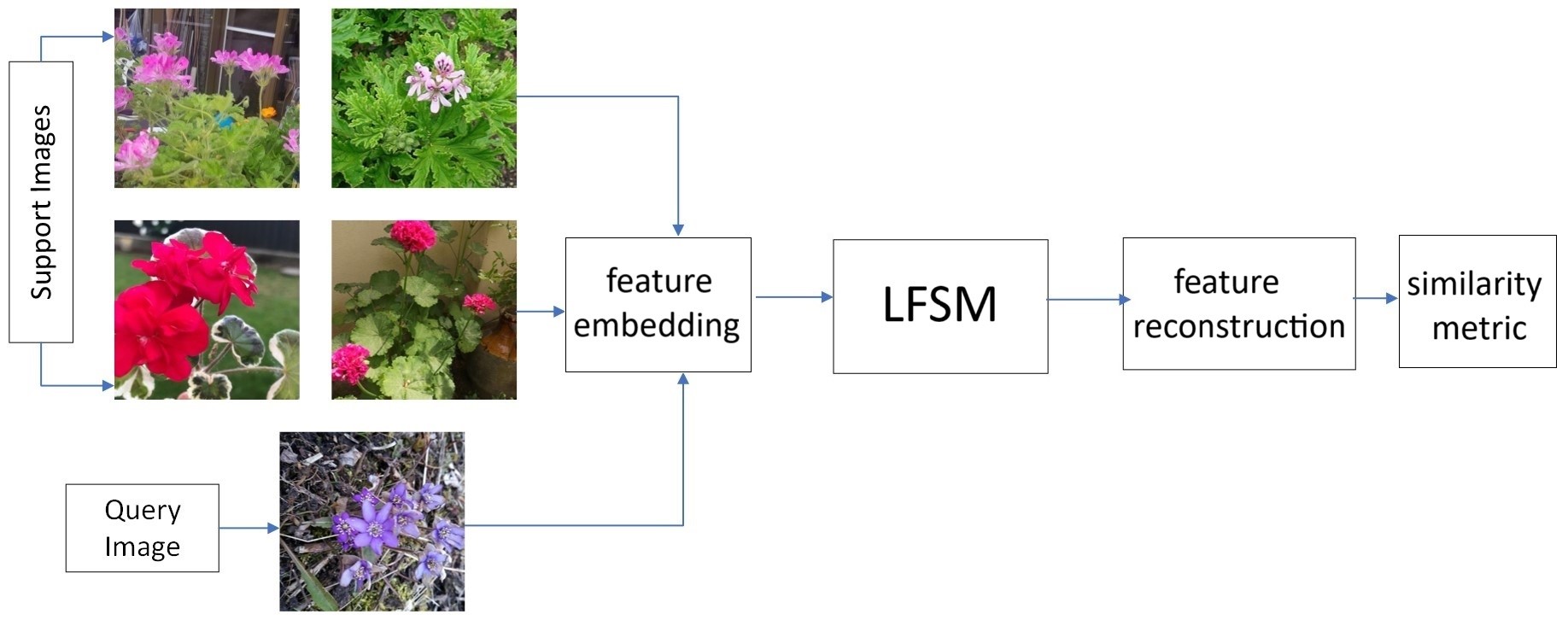, width = 7.5cm}}
  \caption{Local Foreground Selection aware Attentive Feature Reconstruction Network.}
  \label{fig:LFS}
 \end{figure}

The overall framework of our method is illustrated in Figure \ref{fig:LFS} which represents a four step process. For an episode, the support and query samples are passed through feature embedding where feature maps are obtained. To obtain discriminative feature maps suitable for fine grained classification we refine them with our proposed attention mechanism and pass them through the Local Foreground Selection module (LFSM). Subsequently, the output of the LFSM is fed into the feature reconstruction networks of \cite{wu2023bi} and \cite{wertheimer2021few}, which reconstructs query  and support features. Finally, the similarity metric computes the Euclidean distance between the original and reconstructed query and support feature maps. The weighted sum of both distances is used to classify the query image. 
In this section, we present our methodology, beginning with the problem formulation and followed by a detailed description of our method.

\subsection{Problem Formulation}
For a given dataset \textit{D}, it is divided into \textit{$D_{train}$}, \textit{$D_{val}$} and \textit{$D_{test}$} such that the categories in these sets are disjoint. Few-shot classification performs the task of C-way and K-shot classification on \textit{$D_{test}$} by learning the knowledge from \textit{$D_{train}$} and \textit{$D_{val}$}. The task is performed such that C classes from the test set are selected. From each of these classes only K labelled samples are selected which serve as the support samples(\textit{S}) and M unlabelled samples are selected which form part of the query samples(\textit{Q}). In this few-shot task the classification accuracy is determined on \textit{$D_{test}$}.

\begin{figure*}[!htb]
  \centering
   {\epsfig{file = 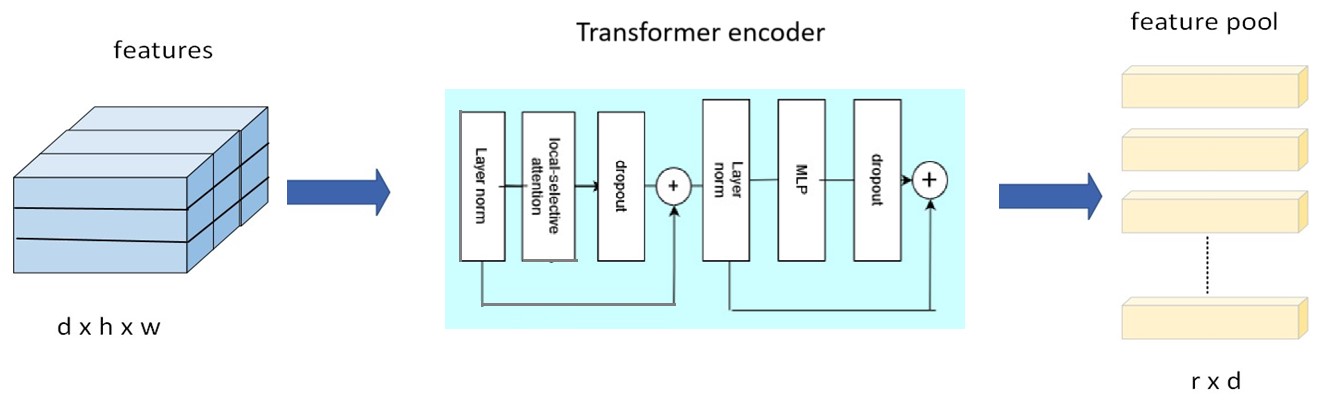, width = 15.5cm}}
  \caption{Local Foreground Selection Module where features are fed into the vision transformer and outputs feature pool.}
  \label{fig:lfsm}
 \end{figure*}

\subsection{Feature Embedding}
The first step in the process is to extract feature maps of  support and query samples by using Convolutional Networks (ConvNet) or Residual Networks (ResNet). Following literature we use the Conv-4 and ResNet-12 networks as backbones to obtain feature maps. The architecture of the Conv-4 and ResNet-12 is the same as in \cite{wertheimer2021few} \cite{wu2023bi}. For a C-way and K-shot task $C \times (K+M)$ images are input to the embedding module where features of each image are extracted.

\subsection{Local Foreground Selection Module(LFSM)}
The LFS module is responsible for generating discriminative features for support and query pools. For a C-way, K-shot task, the extracted features from the embedding module are represented as \textit{$u_{i} = f_{\theta}(u_{i}) \in \mathbb{R}^{d \times h \times w}$}, where \textit{d} denotes the number of channels, \textit{h} is the height of the feature maps, and \textit{w} represents the width. The input features for the Local Foreground Selection Module (LFSM) are denoted as \textit{$u_{i}$}, while the output from this module is \textit{$y_{i} \in \mathbb{R}^{d \times r}$}. 

Within the LFS module, the input features \textit{$u_{i}$} are transformed into \textit{r} local features \textit{$[u_{i}^{1}, u_{i}^{2}, u_{i}^{3}, \ldots, u_{i}^{r}] + E_{pos}$}, which are then fed into a vision transformer encoder.

\begin{figure}[!h]
  \centering
   {\epsfig{file = 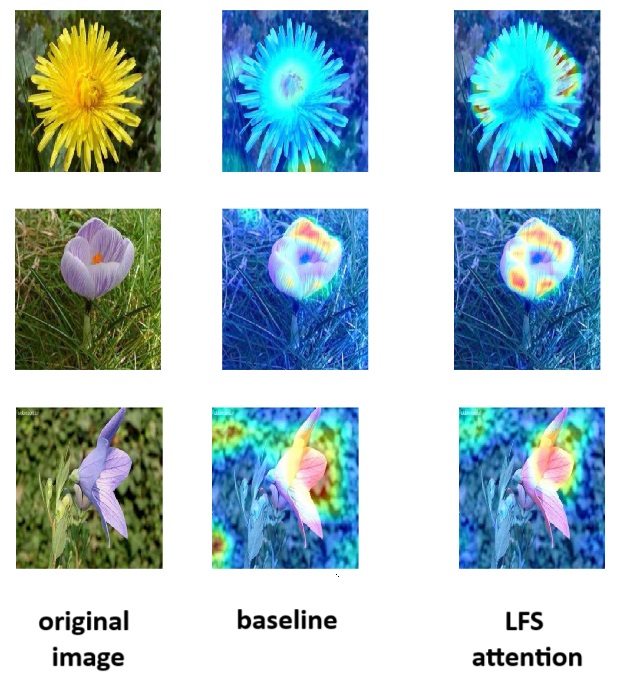, width = 6cm, height = 6 cm}}
  \caption{Heat map comparing the original feature maps to our proposed Local Foreground Selection attention on Oxford Flower-102 dataset.}
  \label{fig:heatmap}
 \end{figure}

The LFSM constructs discriminative feature maps utilizing the proposed Local Foreground Selection attention embedded within a vision transformer encoder. The vision transformer normally employs self-attention but instead we introduce a dual-action attention mechanism. It performs two tasks: the local attention task and foreground selective attention task. The local attention is to emphasize the local details of the plant object. The presence of  natural background in plant images increases intra-class variations and decreases inter-class differences, we aim to mitigate its impact by implementing a foreground selection attention. The foreground selection attention filters out sequences associated with the background of the target plant object, ensuring that only the important tokens relevant to the foreground are retained. The two attentions—local and foreground selection—are aggregated within the vision transformer encoder and generate distinct features for fine grained classification. The architecture of the LFS module is presented in the Figure \ref{fig:lfsm} and a comparison of the heatmap resulting from the LFS attention with the output of the feature embedding is visualized in Figure \ref{fig:heatmap}.

\paragraph{Local Attention}
The local attention mechanism integrates convolutions into the vision transformer block to effectively model local spatial context. Instead of position-wise linear projections for Multi-Head Self-Attention we use convolutional projections for the attention operation. Specifically, depth-wise separable convolutions are utilized for these projections, which are applied to derive the queries (Q), keys (K), and values (V). The feature maps, reshaped earlier in the LFSM module, serve as input for the depth-wise convolutional projections. This function returns the attention scores to the vision transformer encoder.

\paragraph{Foreground Selection Attention}
The purpose of foreground selection attention is to enable a focus on the plant object rather than the background, as it does not contribute to fine-grained classification. In transformer self-attention, all tokens are assigned relevance weights that reduce the influence of less important tokens. Tokens not associated with the plant object receive lower attention scores; however, these background tokens still exert some influence. To further diminish their impact, these tokens are discarded, allowing the feature maps to concentrate on the plant object, thereby aiding in the classification of subtle class differences.

The foreground selection attention generates relevance scores by selecting tokens based on a defined threshold, discarding those below it. The dot product of Q and $K^T$ creates an attention score matrix with dimensions $m \times m$. The relevance of all tokens is then evaluated relative to one another and stored in the relevance matrix, where the relevance of the $i^{th}$ token with respect to the $j^{th}$ token is denoted by the matrix element $relevance_{i,j}$. Relevance scores are calculated for each row of the attention matrix. Tokens representing the plant object yield high relevance scores, while those associated with the background have low scores. Consequently, the relevance scores higher than the threshold are preserved, while lower relevance scores are set to zero. To implement this, we arrange the rows in descending order. After sorting the \(i^{th}\) row of the relevance matrix, we identify the $(m \times \operatorname{FS-ratio})^{th}$ index, where the FS-ratio determines the number of tokens selected and ranges from 0.1 to 1.0. The relevance score at this selected index is recorded as the threshold for comparison. A comparison is then made between $relevance_{ i,j}$ and the threshold; if $relevance_{ i,j}$ exceeds the threshold, it is retained, otherwise set to zero.  The relevance matrix is then normalized by softmax function. The matrix multiplication of the attention score and V is calculated to get the weighted sum over V as indicated by the following equation. This process is mathematically illustrated in the following equations:  

\begin{gather}
\text{relevance} = \frac{Q K^T}{\sqrt{d_k}} \label{eq} \\[1ex]
\text{index} = \operatorname{argsort}\left(-\text{relevance}_{i}\right)\left[\operatorname{FS-ratio} \times m\right] \\[1ex]
\text{relevance}_{i} = \begin{cases}
\text{relevance}_{i}, & \text{if } > \text{relevance}_{i}[\text{index}] \\
0, & \text{otherwise}
\end{cases}
\end{gather}

\begin{equation}
\text{FS-Attention} = \operatorname{softmax}\left(\text{relevance}_{i}\right)  V \label{eq2}
\end{equation}

\indent After getting the Foreground Selection attention(FS-attention) we replace all the non-zero weights to one. In this way the elements of the matrix are either zero or one. The background related tokens are made zero and plant object tokens are made one. This takes the form of an identity matrix in which the relevance score positions that contribute to the object are represented by ones and the non-contributing scores that represent the background are zero.

\[
\text{FS-Attention} = 
\begin{cases}
1 & \text{if } \text{FS-Attention} > 0 \\
0 & \text{otherwise}
\end{cases}
\label{eq3}
\]

\paragraph{Local Foreground Selection Attention}
The local foreground selection attention is a bi-functional attention mechanism designed to capture local details of the plant object, as reflected in the local attention matrix. The foreground selection attention indicates which locations within the attention matrix have scores originating from the plant object versus the background. In the vision transformer encoder, the attention scores derived from local attention are element-wise multiplied with the foreground selection attention matrix. This ensures that regions with lower relevance score corresponding to the background are set to zero, effectively discarding them. Consequently, the resulting attention matrix contains values that specifically pertain to the plant object. Utilizing attention scores from local attention, rather than self-attention, enhances the representation of local details of the plant object. This targeted attention improves focus on the plant object while minimizing background interference, ultimately leading to increased accuracy. 

After getting the attention weights the generated features are obtained by iterative Layer Normalization and Multi Layer Perceptron producing feature pools as output.

\begin{equation}
\hat{y}_{i} = \text{Attention}\left( y_i W^{Q}_{\alpha}, y_i W^{K}_{\alpha}, y_i W^{V}_{\alpha} \right), \quad \hat{y}_{i} \in \mathbb{R}^{r \times d} 
\label{eq4}
\end{equation}

\noindent Where $W^{Q}_{\alpha}$ ,$W^{K}_{\alpha}$ and $W^{V}_{\alpha}$ are learnable weight parameters having size $d \times d$. The $\hat{y}_i$ are obtained using the following equation from Multi Layer Perceptron (MLP) and Layer Normalization (LN).

\begin{equation}
\hat{y}_{i} = MLP \left( LN \left( y_i + \hat{y}_{i} \right) \right)  \label{eq5}
\end{equation}

\section{Experimental Results and Analysis}
In this section the datasets, implementation details and the results are presented.

\begin{table*}[tb]
\vspace{-0.2cm}
\caption{5-way few-shot classification performance for 1-shot and 5-shot on Oxford Flowers-102, iNaturalist 2019, and Plant Net 300-K datasets for plant species with Conv-4 backbone.}
\label{tab:sota1}
\centering
\resizebox{\textwidth}{!}{%
\small 
\begin{tabular}{|p{0.28\textwidth}| p{0.11\textwidth} | p{0.11\textwidth} | p{0.11\textwidth}| p{0.11\textwidth} | p{0.11\textwidth} | p{0.11\textwidth}|} 
\hline
\textbf{Method} & \multicolumn{2}{c|}{\textbf{Oxford Flower-102}}  & \multicolumn{2}{|c|}{\textbf{iNaturalist19}}  & \multicolumn{2}{|c|}{\textbf{Plant Net 300-K}}\\   [8pt]
\hline 
\textbf{Conv-4} & \textbf{1-shot} & \textbf{5-shot} & \textbf{1-shot} & \textbf{5-shot} & \textbf{1-shot} & \textbf{5-shot} \\  [8pt]
\hline
Proto\cite{NIPS2017} & 59.19±.22        & 83.10±.14 & 54.40±.23 & 76.21±.16 & 54.83±.22 & 77.60±.15 \\   [8pt]
DN4\cite{li2019revisiting} & 51.86±.21 & 69.13±.21 & 38.88±.15 & 49.23±.13 & 40.45±.16 & 55.41±.23 \\   [8pt]
CTX\cite{doersch2020crosstransformers} & 70.52±.91 & 80.82±.71 & 46.96±.94 & 63.18±.79 & 49.69±.88 & 65.67±.76 \\  [8pt]
FRN\cite{wertheimer2021few} & 69.40±.20 & 87.45±.12 & 62.08±.22 & 80.19±.15 & 62.01±.22 & 80.81±.14 \\  [8pt]
BiFRN\cite{wu2023bi} & 75.11±.17 & 89.55±.15 & 64.54±.20 & 81.89±.12 & 62.65±.23 & 81.56±.17 \\  [8pt]
LCCRN\cite{li2023locally} & 71.38±.21 & 85.70±.14 & 64.47±.23 & 80.36±.16 & 63.50±.22 & 80.55±.15 \\   [8pt]
BSFA\cite{zha2023boosting} & 71.42±.47 & 83.85±.33 & 60.51±.53 & 75.85±.40 & 62.57±.50 & 77.16±.36 \\  [8pt]
FSCT-Cosine\cite{nguyen2023enhancing} & 66.30±.87 & 82.04±.68 & 51.83±.96 & 64.46±.83 & 52.07±.96 & 66.03±.75 \\   [8pt]
FSCT-Softmax\cite{nguyen2023enhancing} & 62.96±.91 & 76.22±.70 & 50.09±.95 & 62.04±.74 & 49.84±.95 & 62.12±.82 \\   [5pt]
\hline  &&&&&&
\\ [1pt] 
\textbf{Ours: LFS + BiFRN} & 75.71±.10 & \textbf{90.30±.13} & 65.36±.15 & 82.59±.11 & 63.51±.20 & 82.19±.14 \\  [5pt]
\textbf{Ours: LFS + FRN} & \textbf{76.46±.20} & 89.94±.13 & \textbf{66.62±.22} & \textbf{82.76±.15} & \textbf{64.67±.20} & \textbf{82.31±.14} \\  [5pt]
\hline
\end{tabular}
}
\end{table*}

\begin{table*}[tb]
\caption{5-way few-shot classification performance for 1-shot and 5-shot on Oxford Flowers-102, iNaturalist 2019, and Plant Net 300-K datasets for plant species with ResNet-12 backbone.}
\label{tab:sota2}
\centering
\resizebox{\textwidth}{!}{%
\small 
\begin{tabular}{|p{0.24\textwidth}| p{0.11\textwidth} | p{0.11\textwidth} | p{0.11\textwidth}| p{0.11\textwidth} | p{0.11\textwidth} | p{0.11\textwidth}|}
\hline
\textbf{Method} & \multicolumn{2}{|c|}{\textbf{Oxford Flower-102}}  & \multicolumn{2}{|c|}{\textbf{iNaturalist19}}  & \multicolumn{2}{|c|}{\textbf{Plant Net 300-K}}\\ [8pt]
\hline 
\textbf{ResNet-12} & \textbf{1-shot} & \textbf{5-shot} & \textbf{1-shot} & \textbf{5-shot} & \textbf{1-shot} & \textbf{5-shot} \\  [8pt]
\hline
Proto\cite{NIPS2017} & 75.23±.20 & 88.43±.13 & 72.44±.29 & 87.37±.21 & 69.96±.21 & 86.34±.13 \\ [8pt]
DN4\cite{li2019revisiting} & 68.62±.21 & 74.19±.21 & 58.08±.15 &  59.37±.13 & 56.45±.16 & 65.41±.13 \\   [8pt]
CTX\cite{doersch2020crosstransformers} & 77.12±.91 & 83.62±.71 & 55.36±.94 & 60.58±.79 & 59.61±.88 & 70.47±.76 \\  [8pt]
FRN\cite{wertheimer2021few}                  & 79.20±.20 & 90.10±.11 & 74.62±.23 & 87.52±.14 & 72.52±.21 & 86.33±.14 \\ [8pt]
BiFRN\cite{wu2023bi} & 77.75±.17 & 89.94±.15 & 72.24±.16 & 87.20±.12 & 70.84±.15 & 86.52±.12 \\ [8pt]
LCCRN\cite{li2023locally} & 78.95±.19 & 92.60±.10 & 74.96±0.21 & 87.65±.12 & 72.63±.22 & 86.50±.15 \\ [8pt]
BSFA\cite{zha2023boosting} & 77.02±.44 & 89.06±.26 & 76.23±.48 & 87.31±.30 & 74.21±.49 & 86.63±.29 \\ [8pt]
FSCT-Cosine\cite{nguyen2023enhancing} & 63.10±.87 & 85.16±.68 & 58.85±.96 & 61.86±.83 & 62.14±.96 & 71.83±.75 \\   [8pt]
FSCT-Softmax\cite{nguyen2023enhancing} & 59.29±.91 & 79.52±.70 & 59.79±.95 & 58.84±.74 & 60.01±.95 & 67.12±.82 \\   [5pt]
\hline  &&&&&&
\\ [1pt] 
\textbf{Ours: LFS + BiFRN} & 79.77±.12 & 90.87±.10 & 72.95±.15 & 87.67±.11 & 71.04±.15 & 86.70±.14 \\ [5pt]
\textbf{Ours: LFS + FRN} & \textbf{79.85±.20} & \textbf{93.50±.11} & \textbf{79.61±.11} & \textbf{93.33±.15} & \textbf{77.42±.15} & \textbf{88.71±.14} \\ [5pt]
\hline
\end{tabular}
}
\end{table*}

\subsection{Datasets}
The plant species classification problem in this work is considered with the natural background images. We have considered three datasets: one dataset is of flowers only and the other two datasets contain images of different parts of the plant including leaves, branch, flower, fruit, stem, and the whole plant. \textbf{Oxford Flower-102} dataset is a benchmark few-shot fine grained classification dataset \cite{nilsback2008automated} having 102 categories of flowers. \textbf{iNaturalist 2019} dataset is a long-tailed fine grained dataset \cite{van2018inaturalist} having highly similar species. The Plantae category in iNaturalist 2019 dataset contains total 682 species of plants. We have taken a subset of the dataset with 348 species such that each category has at least 50 to maximum 1000 images for our few-shot task. \textbf{Plant Net 300-K} dataset \cite{garcin2021pl} contains 1081 plant species. Also due to its long-tailed nature and for few-shot classification we have considered a subset of 252 species each having 50 to 1000 images. The images of all datasets are resized to 84 x 84. It should be noted that there are no bounding boxes available for the datasets, so the original images are resized to 84 x 84 size without any bounding-box based cropping.

\subsection{Implementation Details}
The experiments are implemented on NVIDIA GeForce RTX 3080 Ti GPU. The implementation of all the methods is performed using Pytorch. The experiments are conducted on Conv-4 and ResNet-12 backbones and their architectures are the same as in \cite{wertheimer2021few}. All methods including the baselines, state-of-the art and ours are trained from scratch. The training is performed for 1200 epochs using SGD with Nesterov momentum of 0.9. The initial learning rate is taken as 0.1 and weight decay is 5e-4. Learning rate decreases by a scaling factor of 10 every 400 epochs. For Conv-4 models, we train using 30-way 5-shot episodes, and test for 1-shot and 5-shot episodes. The query images per class for Conv-4 are 15 for both 1-shot and 5-shot. For ResNet-12 models the training episodes are 15-way 5-shot. Some data augmentation methods like centre crop, random horizontal flip and colour jitter are used. The best-performing model are selected based on the validation set and are validated every 20 epochs.  For all experiments, we report the mean accuracy of 10,000 randomly generated tasks on $\it{D_test}$ with $95\%$ confidence intervals on the standard 5-way, 1-shot and 5-shot settings.

\subsection{Comparison with State-of-the-Art}
The accuracy of our method is tested for plant species classification via few-shot learning. Experiments are conducted on the above mentioned three plant species datasets. The state-of-the art methods have worked on other fine-grained datasets but not on the aforementioned plant species datasets. So, we obtain results on these methods ourselves on the official codes provided for each work. The backbones used in these comparative methods are Conv-4 and ResNet-12 which test the 5-way 5-shot and 5-way 1-shot performance for plant species classification. The state-of-the-art methods that we compare our method to include  ProtoNet \cite{NIPS2017}, DN4 \cite{li2019revisiting}, CTX \cite{doersch2020crosstransformers}, FRN \cite{wertheimer2021few}, LCCRN \cite{li2023locally}, BiFRN \cite{wu2023bi}, BSFA \cite{zha2023boosting} and FSCT \cite{nguyen2023enhancing}.   \\

\indent A comparison of the performance of our method with the state-of-the-art methods is reflected in Table \ref{tab:sota1} for Conv-4 backbone and Table \ref{tab:sota2} for ResNet-12 backbone. For the Conv-4 backbone our method outperforms all other state of the art methods for all three plant species dataset. If we carefully observe  we can see that the performance of our method with FRN \cite{wertheimer2021few} significantly improves from 2-7 \%. The 1-shot performance improves significantly for all three datasets. The accuracy also increases appreciably for 5-shot as well. When our method is used with BiFRN \cite{wu2023bi} our local foreground selection attention works well for both 1-shot and 5-shot as it removes the effect of background and focus on local details. 
For the ResNet-12 backbone also the performance of our method is better for 5-way 5-shot and 5-way 1-shot  for all three datasets. The accuracy improves by 1-6 \% when our attention mechanism is introduced with FRN and BiFRN. The LFS attention used with FRN outperforms all other methods.   
The FRN method complemented with our method outperforms all state of the arts as well as the LFS with BiFRN. This show that our attention mechanism gives highly discriminative feature maps that are suitable for feature reconstruction. Even though FRN performs unidirectional feature reconstruction the LFS attention make its performance comparable to the bi-directional feature reconstruction in BiFRN.

\subsection{Ablation Study}
The ablation study performed on our proposed method justify the design choices made in this model. The ablation study reflects on how accuracy behaves by adding and removing the components and changing the parameters we have proposed in our method. 

\paragraph{The Effectiveness of Local and Foreground Selection Attention}

Our proposed local foreground selection attention improves the overall accuracy when used with both FRN and BiFRN. So, we test the effectiveness of our local foreground selection attention as compared to self attention, local attention and selective attention . A comparison is presented with the baseline i.e ProtoNet \cite{NIPS2017} and other aforementioned attentions . Our local foreground selection attention(LFS-attention) performs better with the combination of both local and selective attention. They do not give the best results when used alone which is the evidence that LFS-attention is more effective. The results are shown in Table \ref{tab:ablation1} and comparison is made for two datasets of plant species considered in this paper. The results reflect that our proposed LFS-attention is a superior approach for fine grained few-shot classification.

\begin{table*}[tb]
\caption{Ablation study for 5-way few-shot classification performance for 1-shot and 5-shot on Oxford Flowers102, iNaturalist 2019, and Plant Net 300-K datasets for plant species with ResNet-12 backbone.}
\label{tab:ablation1}
\centering
\resizebox{\textwidth}{!}{%
\footnotesize 
\begin{tabular}{|p{0.09\textwidth}|p{0.15\textwidth}|p{0.09\textwidth}|p{0.09\textwidth}|p{0.09\textwidth}| p{0.09\textwidth}|}
\hline
\multirow{2}{*}{\textbf{Backbone}} & \multirow{2}{*}{\textbf{Method}} & \multicolumn{2}{c|}{\textbf{Oxford Flower-102}} & \multicolumn{2}{c|}{\textbf{iNaturalist19}} \\ 
\cline{3-6}
 &  & 1-shot & 5-shot & 1-shot & 5-shot \\ 
\hline
Conv-4 & Baseline & 59.19±.22 & 83.10±.14 & 54.40±.22 & 76.21±.17 \\
\hline
\multirow{4}{*}{\parbox{2cm}{FRN \\ (Conv-4)}} 
& Self-attention & 74.20±.20 & 87.96±.13  & 64.21±.21 & 81.21±.14 \\ 
& Local attention & 74.89±.18 & 88.20±.13 & 64.68±.25 & 81.51±.11 \\ 
& Selective attention & 75.15±.17 & 88.65±.14 & 64.92±.20 & 81.70±.12 \\ 
& \textbf{LFS-attention} & \textbf{76.46±.20} & \textbf{89.94±.13} & \textbf{66.62±.22} & \textbf{82.76±.15} \\ 
\hline
\multirow{4}{*}{\parbox{2cm}{BiFRN \\ (Conv-4)}} 
& Self-attention & 75.11±.17 & 89.55±.15 & 64.54±.20 & 81.89±.12 \\ 
& Local attention & 74.91±.17 & 89.60±.15 & 65.03±.20 & 82.28±.12 \\ 
& Selective attention & 74.84±.17 & 89.79±.15 & 64.91±.20 & 82.36±.12 \\ 
& \textbf{LFS-attention} & \textbf{75.71±.10} & \textbf{90.30±.13} & \textbf{65.36±.15} & \textbf{82.59±.11} \\ 
\hline
ResNet-12 & Baseline & 70.99±.20 & 86.99±.13 & 64.59±.29 & 81.65±.21 \\ 
\hline
\multirow{4}{*}{\parbox{2cm}{FRN \\ (ResNet-12)}} 
& Self-attention & 79.01±.20 & 91.50±.11 & 77.69±.11  &  90.52±.16 \\ 
& Local attention & 78.88±.16 & 91.89±.12 & 78.12±0.15  & 91.26±.13 \\ 
& Selective attention & 78.95±.21 & 92.26±.15 & 78.54±.12 & 91.98±.17  \\ 
& \textbf{LFS-attention} & \textbf{79.85±.20} & \textbf{93.50±.11} & \textbf{79.61±.11}  &\textbf{93.33±.15} \\ 
\hline
\multirow{4}{*}{\parbox{2cm}{BiFRN \\ (ResNet-12)}}
& Self-attention & 77.75±.17 & 89.94±.15 & 72.24±.16 & 87.20±.12 \\ 
& Local attention & 78.31±.17 & 90.08±.15 & 71.97±.16 & 87.12±.12 \\ 
& Selective attention & 77.75±.28 & 88.79±.15 & 72.44±.16 & 87.27±.12 \\ 
& \textbf{LFS-attention} & \textbf{79.77±.12} & \textbf{90.87±.10} & \textbf{72.95±.15} & \textbf{87.67±.11} \\ 
\hline
\end{tabular}
}
\end{table*}

\paragraph{The Effect of FS-ratio}
$\operatorname{FS-ratio}$ defined in the foreground selection attention part defines the number of tokens that will be accepted. By incorporating FS-ratio we can see the impact of number of tokens that will be accepted to provide the best results. The accuracy of classification based on different $\operatorname{FS-ratio}$ varies according to the backbone and the dataset. This is reflected in the Table \ref{tab:ab21} and Table \ref{tab:ab22}. This presents evidence for the effect of number of background tokens accepted in the local foreground selection attention. 
The results in the tables present the performance on different select ratios. While the $\operatorname{FS-ratio}$ may range from 0.1 to 1.0. The $\operatorname{FS-ratio}$ that gives the best performance is highlighted with bold letters,  for each method and the respective dataset. 
For Oxford Flower-102 dataset the best results are given by FS-ratio 0.3 for FRN(Conv-4 backbone) and 0.5 for BiFRN(Conv-4 backbone. For ResNet-12 backbone the FS-ratio 0.1 gives best results for both FRN and BiFRN methods.
For the iNaturalist 2019 dataset the Conv-4 backbone gives best results with FS-ratio 0.3 for FRN and 0.5 for BiFRN. While for ResNet-12 backbone 0.3 FS-ratio works best for both FRN and BiFRN. 
In this way different datasets have different suitable FS-ratio for each method and the relevant backbone.

\begin{table}[!h]
\caption{Performance with different select ratios for Oxford flowers 102 dataset for 5-way 5-shot and 5-way 1-shot on Conv-4 and ResNet-12 backbones}
\label{tab:ab21}
\centering
\begin{tabular}{|p{0.075\textwidth}|  p{0.045\textwidth} | p{0.045\textwidth} | p{0.045\textwidth} | p{0.045\textwidth} | p{0.045\textwidth} |}
\hline
\textbf{FRN }  & \multicolumn{5}{|c|}{\textbf{ $\operatorname{\textbf{FS-ratio/ Oxford flower-102}}$}}        \\
\hline 
   & \textbf{0.1} & \textbf{0.3} & \textbf{0.5} & \textbf{0.7} & \textbf{0.9}      \\
\hline
Conv (5-shot) &  89.42 & \textbf{89.94} & 88.83 & 89.30 & 88.70     \\
Conv (1-shot) & 76.03 & \textbf{76.46} & 75.26 & 75.16 & 74.20     \\
\hline
ResNet (5-shot) & \textbf{93.50} & 93.37 & 93.04 & 92.88 & 93.12 \\
ResNet (1-shot) & \textbf{79.85} & 79.73 & 79.43 & 79.46 & 79.45   \\
\hline
\textbf{BiFRN }  & \multicolumn{5}{|c|}{\textbf{ $\operatorname{\textbf{FS-ratio/ Oxford flower-102}}$}}        \\
\hline 
   & \textbf{0.1} & \textbf{0.3} & \textbf{0.5} & \textbf{0.7} & \textbf{0.9}      \\
\hline
Conv (5-shot) &  89.69 & 89.32 & \textbf{90.30} & 89.44 & 89.91     \\
Conv (1-shot) & 74.96 & 74.64 & \textbf{75.71} & 74.51 & 74.91     \\
\hline
ResNet (5-shot) & \textbf{90.87} & 90.40 & 89.90 & 89.81 & 90.24 \\
ResNet (1-shot) & \textbf{79.77} & 78.97 & 77.83 & 78.55 & 78.51   \\
\hline
\end{tabular}
\label{tab4}
\end{table}

\begin{table}[!h]
\caption{Performance with different select ratios for iNaturalist 19 dataset for 5-way 5-shot and 5-way 1-shot on Conv-4 and ResNet-12 backbones}
\label{tab:ab22}
\centering
\begin{tabular}{|p{0.075\textwidth}| p{0.045\textwidth}| p{0.045\textwidth} | p{0.045\textwidth} | p{0.045\textwidth}| p{0.045\textwidth} |}
\hline
\textbf{ FRN}  & \multicolumn{5}{|c|}{\textbf{$\operatorname{\textbf{FS-ratio/ iNaturalist 2019}}$}}        \\
\hline 
   & \textbf{0.1} & \textbf{0.3} & \textbf{0.5} & \textbf{0.7} & \textbf{0.9}      \\
\hline
Conv (5-shot) &  82.69 & \textbf{82.76} & 82.48 & 82.52 & 82.40     \\
Conv (1-shot) & 66.15 & \textbf{66.62} & 66.14 & 65.83 & 66.01     \\
\hline
ResNet (5-shot) & 93.21 & \textbf{93.33} & 92.82 & 93.06 & 92.82 \\
ResNet (1-shot) & 79.44 & \textbf{79.61} & 78.81 & 79.12 & 78.69   \\
\hline
\textbf{BiFRN }  & \multicolumn{5}{|c|}{\textbf{$\operatorname{\textbf{FS-ratio/ iNaturalist 2019}}$}}        \\
\hline 
   & \textbf{0.1} & \textbf{0.3} & \textbf{0.5} & \textbf{0.7} & \textbf{0.9}      \\
\hline
Conv (5-shot) &  82.39 & 82.05 & \textbf{82.59} & 81.82 & 82.50     \\
Conv (1-shot) & 65.00 & 64.43 & \textbf{65.36} & 64.63 & 65.18     \\
\hline
ResNet (5-shot) & 86.68 & \textbf{87.67} & 86.59 & 87.37 & 85.92 \\
ResNet (1-shot) & 70.99 & \textbf{72.95} & 70.98 & 72.19 & 70.25   \\
\hline
\end{tabular}
\end{table}

\section{\uppercase{Conclusion}}
\label{sec:conclusion}
In this paper we propose a novel Local Foreground Selection(LFS) attention which is a tailored module to generate discriminative query and support features suitable for few-shot fine grained classification. The task of plant species classification has large intra-class and small inter-class variations which need to be optimized. The Foreground Selection attention works by highlighting the foreground and reducing the effect of background which contributes to alleviate the high intra-class variations. Secondly, we extract local spatial details to focus on the fine-grained details of the plant object with the local attention. Combining the local and foreground selection attentions optimizes both inter-class and intra-class differences. Experiments conducted with the proposed method show that it improves the classification accuracy on three plant species datasets and the ablation studies validate the effectiveness of LFS attention.

\bibliographystyle{apalike}
{\small
\bibliography{example}}

\end{document}